\documentclass[conference]{IEEEtran}
\IEEEoverridecommandlockouts

\usepackage{cite}
\usepackage{amsmath,amssymb,amsfonts}
\usepackage{graphicx}
\usepackage{textcomp}
\usepackage{color}
\usepackage{url}
\providecommand{\href}[2]{#2 (\url{#1})}
\usepackage{algorithm}
\usepackage{algorithmic}

\begin{document}

\title{Quantum Computing Based Image Denoising
\thanks{Corresponding author: Taranum Wasu, taranumwasu121@gmail.com. \copyright\ 2026 IEEE. Personal use of this material is permitted. Permission from IEEE must be obtained for all other uses, in any current or future media, including reprinting/republishing this material for advertising or promotional purposes, creating new collective works, for resale or redistribution to servers or lists, or reuse of any copyrighted component of this work in other works.}
}

\author{

\IEEEauthorblockN{1\textsuperscript{st} Taranum Wasu}
Research Student \\
\IEEEauthorblockA{\textit{Thapar University} \\ 
Patiala, India \\
taranumwasu121@gmail.com}

\and

\IEEEauthorblockN{2\textsuperscript{nd} Harinder Kaur}
\IEEEauthorblockA{Department of Mathematics\\
\textit{Maharaja Ranjit Singh Polytechnic University}\\ 
Bathinda, India \\
h.kaur1989@gmail.com}

\and

\IEEEauthorblockN{3\textsuperscript{rd} HS Pannu}
\IEEEauthorblockA{Department of Computer Science\\ 
\textit{Thapar University}\\
Patiala, India \\
hspannu@thapar.edu}

}

\maketitle

\begin{abstract}
Image denoising aims to remove noise while preserving structural details and remains a challenging problem in image processing. Existing methods often struggle to balance noise suppression and detail preservation. In this paper, we propose a hybrid quantum--classical image denoising framework based on variational quantum circuits. The input image is divided into $8\times8$ patches, encoded into quantum states, and processed using a variational quantum circuit to estimate noise components. The denoised image is reconstructed by aggregating processed patches. The proposed approach is evaluated using Peak Signal-to-Noise Ratio (PSNR), Structural Similarity Index (SSIM), and entropy. In a ten-image evaluation using the same clean image content for training and testing with newly generated noise, the complete pipeline achieves higher mean PSNR and SSIM than Gaussian, median, bilateral, and a small CNN baseline. Per-image SSIM is mixed. A second evaluation on 14 held-out BSD68 images places the pipeline behind BM3D and NLM but ahead of classical filtering baselines at low and moderate noise. These results are a limited proof of concept rather than evidence of generalization or quantum advantage.
\end{abstract}


\section{Introduction}
Digital images have become an integral part of our lives, providing reliable information for security systems, satellite imaging, medical sciences and social media \cite{acharya2005image}. However, images often suffer from artefacts and noise during acquisition or processing. Artifacts are structured (such as rings or lines) while noise is random, but both are unwanted additions that degrade image quality and hinder information retrieval in image processing and computer vision tasks. The task of image denoising is to remove such unwanted components while preserving the important features of the image. Since digital images are stored as matrices, matrix computation enables quality enhancement through pixel-by-pixel spatial analysis, and the frequency domain allows analysis of the rate of pixel changes across the image. In the spatial domain, techniques such as mean and Gaussian filtering apply a small kernel that slides over the image, modifying pixel values through weighted sums to sharpen or smooth the image, and thresholding filters can remove specific ranges of values; for example, salt-and-pepper noise can be replaced by neighbourhood averages. In the frequency domain, transforms such as Fourier and Laplace can generate a cleaner signal. These methods are effective to some extent but often require a trade-off between preserving fine details and removing noise.

For colored or high-definition images, traditional image processing methods often face computation and speed limitations. Quantum computing (QC) has emerged as a strategy to deal with larger datasets \cite{wang2022review}. QC replaces the bit with a qubit, which has two key properties: (a) superposition, meaning a qubit can be both 0 and 1 simultaneously, and (b) entanglement, meaning qubits can correlate with each other regardless of distance. These follow the principles of quantum physics, analogous to the behaviour of atomic and sub-atomic particles. The term QC was popularized by Nobel laureate Richard Feynman in the early 1980s.

Recently, quantum image processing (QIP) has suggested flexible representations of images as quantum states to enable parallel computing and probability-based noise reduction. Using the Python-based SDK Qiskit, QIP has become feasible on classical computers \cite{boreiri2022optimized}, and image denoising using QC is an emerging area for large-scale image and video applications. However, QIP is slow on current computers because they must simulate quantum behaviour: $n$ qubits require $2^n$ probability states, and superposition and entanglement must be stored explicitly rather than existing simultaneously as on real quantum hardware. This leads to exponential computational overhead. To mitigate this, we divide the image into $8\times8$ blocks, denoise each block, and combine them to form the final image.
Next, we discuss state-of-the-art denoising techniques, followed by the proposed technique, results, and conclusion.

\section{Literature Review}
Gaussian noise is random and has a similar nature to the bell-shaped probability distribution. Nearly each pixel of the image gets affected and the reasons can be thermal, amplifier, or electronic sensors. This type of noise has a grainy nature and is just like the bell curve, where most of the noise values are closer to the mean value of the noise. Detailed survey can be found in \cite{mafi2019comprehensive}. Image denoising in deep learning has been surveyed in \cite{jebur2024comprehensive}, and its challenges have been studied in \cite{li2023ntire}. Several classical image denoising techniques were considered as baselines in this work. Gaussian filtering smooths images by applying a weighted averaging kernel based on a Gaussian distribution, which helps remove Gaussian noise but may blur edges.

Median filtering replaces each pixel value with the median value of its neighbourhood and is particularly effective for removing impulse noise such as salt-and-pepper noise. Bilateral filtering performs edge-preserving smoothing by combining spatial proximity and pixel intensity similarity. In \cite{omer2018denoising} CT images were used to denoise the salt and pepper noise using various median filter algorithms. 

A bilateral filter (BF) operates on the pixels by finding the averages but does not disturb the edges. But parameters of the bilateral filters needs to be tuned since it affects the denoising results. In \cite{zhang2008multiresolution} the optimal parameters selection along with multiresolution bilateral filter extension has been studied. So BF is applied to the subband approximations of the decomposition of the signal using the wavelents. In \cite{elhoseny2019optimal} bio-inspired filtering using optimizations has been considered for medical imaging. The Gaussian and spatial weight parameters for BF has been searched using  DragonFly and Modified Firefly algorithms.  

Convolutional neural network (CNN) are also popular to remove various types of noises such as salt-and-pepper, Gaussian, speckle, impulse etc.   Various CNN based algorithms have been studied in \cite{ilesanmi2021methods}. Deep learning approaches such as DnCNN \cite{zhang2017dncnn} and FFDNet \cite{zhang2018ffdnet} employ residual learning and flexible noise-level handling to achieve strong denoising performance. Among classical methods, Block-Matching and 3D filtering (BM3D) \cite{dabov2007bm3d} is widely regarded as a state-of-the-art (SOTA) baseline; it groups similar image patches and performs collaborative filtering in a sparse transform domain, and provides relevant context for future comparisons. BM3D is not part of the five-method experiment reported here.
Quantum and deep convolution generative adversarial networks have also been studied recently for medical image denoising in \cite{nandal2025image}, where a QC layer is added to a deep CNN architecture to improve PSNR and SSIM.

\section{Methodology}
The proposed Hybrid Quantum Model (HQM) combines classical preprocessing with a variational quantum circuit. Classical computing handles image pre-processing and representation, splitting the image into smaller blocks that can be computed quickly, while QC performs the transformations that extract features and estimate noise. For the experiments, we use images from various datasets and introduce Gaussian noise, as explained next.

\subsection{Gaussian Noise}
Gaussian noise is often used to generate noisy images because it can closely describe the natural noise in the nature for real life images. Many small random noises gets combined to form Gaussian in nature; Gaussian noise is simple to generate, well defined mathematically and it is also a standard noise to test and compare with the state-of-art benchmarks \cite{sun2024learning}. Thus it is easy to work with for the experiments and also behaves like real life disturbances in the signals and imaging systems, which makes it a popular choice. 
Gaussian noise is added to simulate real-world imaging conditions. The noisy image $I_n(x,y)$ is generated as:
\begin{equation}
I_n(x,y) = I(x,y) + n(x,y)
\end{equation}
where $n(x,y)$ represents Gaussian noise with zero mean $\mu=0$  and standard deviation $\sigma$. In our experiments the noise standard deviation is set to $\sigma=0.1$ on images normalized to the range $[0,1]$. Gaussian noise introduces random variations in pixel intensities, which appear as grainy patterns 
in the image \cite{oravec2011efficiency}.

Now we need to divide the image into patches of $8\times8$ so that quantum computing can denoise each patch quickly and then patches are connected again. 

\subsection{Image into Patches}
\begin{figure}[htbp]
\centering
\includegraphics[width=0.72\columnwidth]{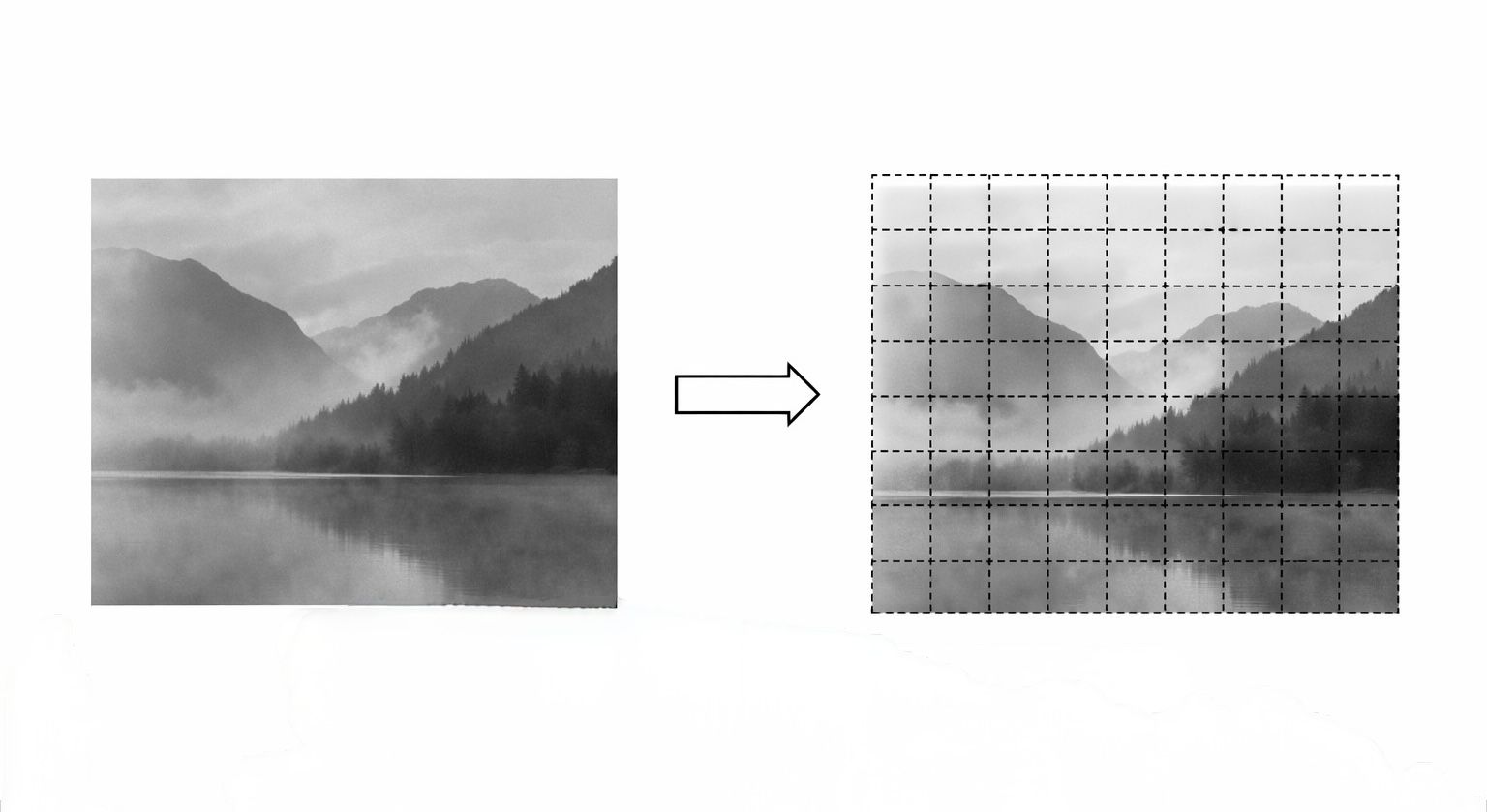}
\caption{Patch-based decomposition of the input image into $8\times8$ blocks, enabling efficient quantum processing of local regions.}
\label{fig_patch}
\end{figure}

Large images are divided into smaller patches of size $8\times8$ pixels \cite{kumar2024quantum} as follows:
\begin{equation}
P_k = I(x:x+7, y:y+7)
\end{equation}
where $P_k$ represents the $k^{th}$ patch.
Patch-based processing is illustrated in Fig.~\ref{fig_patch}. The input image is divided into smaller $8\times8$ blocks which are processed independently. This significantly reduces the 
computational complexity when simulating quantum circuits on classical hardware.
After the patch division of an image, each patch is denoised using QIP as follows.
\subsection{Hybrid Quantum Denoising Model}

The proposed method is summarized in Algorithm 1, following the pipeline: Noisy Image $\rightarrow$ Patch $\rightarrow$ Quantum Encoding $\rightarrow$ VQC $\rightarrow$ Measurement Map $\rightarrow$ Predicted Noise $\rightarrow$ Final Denoised Image $\rightarrow$ \{PSNR, SSIM, Entropy\}. A Variational Quantum Circuit (VQC) is a quantum circuit, analogous to a neural network, whose gate parameters are trainable; unlike a standard circuit with fixed gates, these parameters are optimized during computation. Each $8\times8$ patch $x_i$ is converted into quantum rotation angles $\theta_i$, processed by the VQC $U_{vqc}$, measured to obtain features $z_{i,j}$, used to predict the noise $N_i$, which is then subtracted to obtain the denoised patch.

The implemented VQC uses two trainable RX-rotation layers with CNOT entanglement. The RY gates encode the six features produced by the classical input layer. The full trainable model predicts a residual noise patch.

\begin{figure}[htbp]
\centering
\includegraphics[width=0.55\columnwidth]{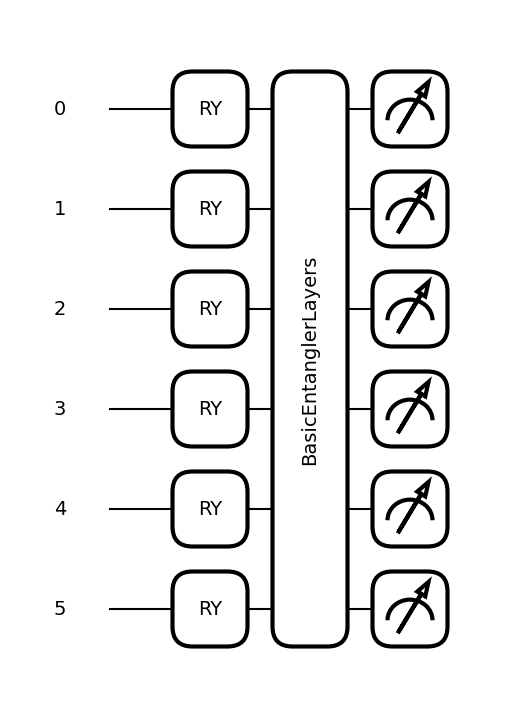}
\caption{
Variational circuit of the proposed HQM. Each of the six qubits receives one encoded rotation angle through an RY gate, followed by two basic entangler layers (trainable RX rotations and a ring of CNOT gates). The Pauli-Z expectation value of every qubit is measured and passed to the classical output layer.}
\label{fig_circuit}
\end{figure}

The variational circuit structure used for all experiments is shown in Fig.~\ref{fig_circuit}. 

Dirac notation \cite{hughes2021qubit} is used to express quantum states of $i^{th}$ qubit. Let $S_i$ be the quantum state of $i^{th}$ qubit, $a_i$ is the probability amplitude of state $|0\rangle$, $b_i$ is the probability amplitude of state $ |1\rangle$. Then for a single $i^{th}$ qubit the quantum state is defined as
\begin{equation}
|S_i\rangle = a_i |0\rangle + b_i |1\rangle
\end{equation}
The additional normalized condition is 
\begin{equation}
|a_i|^2 + |b_i|^2 = 1
\end{equation}
where $|a_i|^2$ and $|b_i|^2$ are probabilities of qubit being 0 and 1 respectively. The vector form can be written as 
\begin{equation}
|S_i\rangle =
\begin{bmatrix}
a_i \\
b_i
\end{bmatrix}
\end{equation}
Table 1 summarizes the notation used in the proposed hybrid quantum--classical image denoising framework.
\begin{table}[t]
\centering
\footnotesize
\caption{Notation Used in the Proposed Quantum Image Denoising Model}
\begin{tabular}{|c|l|}
\hline
Symbol & Description \\
\hline
$I$ & Original image \\
$Y$ & Noisy image ($Y = I + N$) \\
$N$ & Noise in the image \\
$x_i$ & $i^{th}$ image patch extracted from $Y$ \\
$M$ & Number of image patches \\
$d$ & Dimension of flattened patch \\
$f_{enc}$ & Classical feature encoder \\
$\theta_i$ & Rotation angles generated from patch $x_i$ \\
$|\psi(\theta_i)\rangle$ & Quantum state after encoding the patch \\
$U_{enc}$ & Quantum encoding operation \\
$U_{vqc}(\alpha)$ & VQC with parameters $\alpha$ \\
$\alpha$ & Trainable parameters of QC \\
$|\phi(\theta_i,\alpha)\rangle$ & Quantum state after VQC processing \\
$Z_j$ & Pauli-Z measurement operator on qubit $j$ \\
$z_{i,j}$ & Measurement expectation value from qubit $j$ \\
$z_i$ & Measurement feature vector from all qubits \\
$f_{pred}$ & Classical noise prediction function \\
$\hat{N}_i$ & Predicted noise in patch $x_i$ \\
$\hat{x}_i$ & Denoised patch \\
$\hat{I}$ & Final reconstructed denoised image \\
$\mathcal{R}$ & Patch reconstruction operator \\
$\mathcal{L}$ & Training loss function \\
\hline
\end{tabular}
\end{table}
After processing all patches independently, they are merged together to reconstruct the final denoised image. 
The reconstructed image is post-filtered as described in the experimental setup.
\begin{equation}
\hat{I} = \operatorname{clip}\!\left(B(\mathcal{R}(\{\hat{x}_i\})),0,1\right),
\end{equation}
where $B$ denotes the bilateral post-filter.

\begin{algorithm}
\caption{Proposed Hybrid Algorithm for Image Denoising (Classical + Quantum)}
\begin{algorithmic}[1]

\REQUIRE Noisy image $Y$, trained parameters $\Theta$
\ENSURE Denoised image $\hat{I}$

\STATE Divide $Y$ into $M$ overlapping patches $\{x_i\}_{i=1}^{M}$

\FOR{$i = 1$ to $M$}

\STATE Encode classical features  
$\theta_i = f_{\mathrm{enc}}(x_i)$

\STATE Prepare quantum state  
$|\psi(\theta_i)\rangle = U_{\mathrm{enc}}(\theta_i)|0\rangle^{\otimes n}$

\STATE Apply VQC  
$|\phi(\theta_i,\alpha)\rangle = U_{\mathrm{vqc}}(\alpha)|\psi(\theta_i)\rangle$

\STATE  Expectation values  
$z_{i,j} = \langle \phi(\theta_i,\alpha)|Z_j|\phi(\theta_i,\alpha)\rangle$

\STATE Predict noise  
$\hat{N}_i = f_{\mathrm{pred}}(z_i)$

\STATE Obtain denoised patch  
$\hat{x}_i = \operatorname{clip}(x_i - \hat{N}_i,0,1)$

\ENDFOR

\STATE Reconstruct denoised image  
$\hat{I} = \mathcal{R}(\{\hat{x}_i\}_{i=1}^{M})$

\STATE Apply the bilateral post-filter to $\hat{I}$ and clip to $[0,1]$
\RETURN $\hat{I}$

\end{algorithmic}
\end{algorithm}

\section{Dataset}

\begin{figure}[htbp]
\centering
\includegraphics[width=0.7\columnwidth]{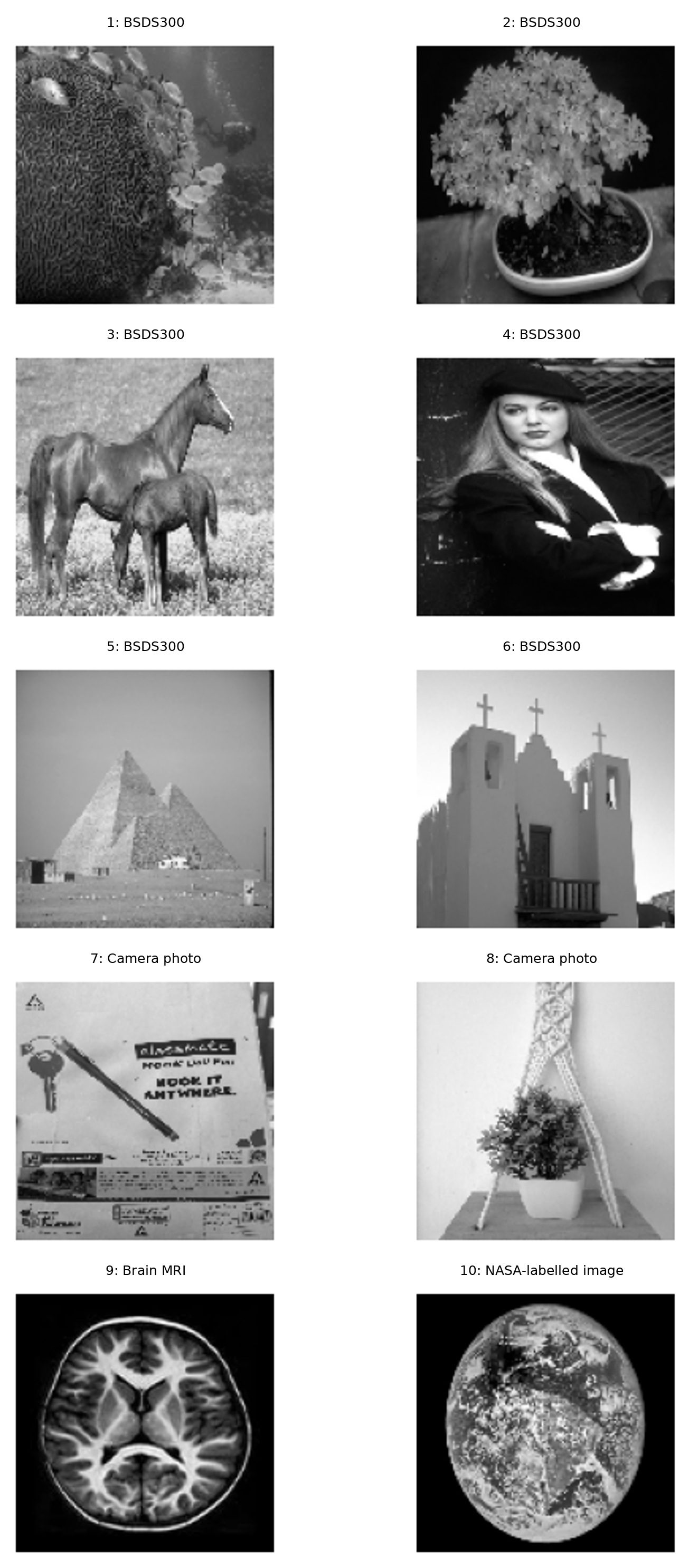}
\caption{Sample images from the dataset used in this study. The dataset consists of 10 grayscale 
images: six natural scenes from BSDS300, two camera photos, one brain MRI scan, and one NASA-labelled Earth image. The panels follow the image order used in Tables~\ref{tab_psnr}--\ref{tab_entropy}.}
\label{fig_dataset}
\end{figure}

The dataset used in this study consists of ten grayscale images resized to $128\times128$ pixels: six natural images from the Berkeley Segmentation Dataset (BSDS300), two consumer camera photos, one brain MRI scan, and one Earth image labelled as NASA imagery in the experiment notebook. Gaussian noise with standard deviation $0.1$ (on pixel values in $[0,1]$) was added to each image, and the noisy images were clipped to the valid range before denoising. All methods were evaluated on the same noisy images. A snapshot of the dataset is shown in Fig.~\ref{fig_dataset}.

%
%
\subsection{Experimental Setup}
All experiments were conducted on a classical computer using the PennyLane library (default.qubit simulator) with a PyTorch interface. The variational circuit acts on 6~qubits. Each $8\times8$ image patch is flattened to a 64-dimensional vector and mapped by a classical linear layer ($64\rightarrow6$) with a tanh activation to rotation angles, which are encoded onto the qubits through RY gates. The trainable circuit consists of two basic entangler layers (trainable RX rotations followed by a ring of CNOT gates), and the Pauli-Z expectation value of each qubit is measured. The measurement vector passes through a ReLU activation and a second linear layer ($6\rightarrow64$) that maps it to the predicted noise patch. The denoised patch is obtained by subtracting the predicted noise from the input patch and clipping to $[0,1]$.

For the ten-image study, patches are extracted with a stride of 6, so neighbouring patches overlap, and the final image is reconstructed by averaging the overlapping regions. A bilateral filter (diameter 5, sigma values 75) is then applied to the reconstructed image as a post-processing step. The model is trained with the Adam optimizer (learning rate 0.01) to minimize the mean squared error between denoised and clean patches, for 5~epochs over 80 randomly sampled patches per image. NumPy, PyTorch, and Python random seeds were initialized to 42. Both learned models used the same ten clean images for training and evaluation, with fresh noise generated during evaluation; no held-out image split was used. The CNN baseline is a two-layer convolutional network trained with Adam (learning rate 0.001) for 5~epochs on the same images; the classical baselines are Gaussian filtering ($5\times5$ kernel), median filtering ($3\times3$) and bilateral filtering (diameter 9, sigma values 75). The extended BSD68 evaluation (Section~\ref{sec_bsd68}) uses the separate protocol and training settings described there.

\subsection{Evaluation Metrics}
To quantitatively evaluate the performance of the proposed denoising method, we have chosen 3 commonly used image quality metrics: Peak Signal-to-Noise Ratio (PSNR), Structural Similarity Index Measure (SSIM), and entropy. These metrics provide complementary information about reconstruction quality, structural preservation, and information content in the denoised images.

Peak Signal-to-Noise Ratio (PSNR) quantifies reconstruction fidelity as the ratio between the peak signal power and the error introduced during restoration, derived from the mean squared error (MSE); higher values indicate better noise removal.
\begin{equation}
PSNR = 10\log_{10}\left(\frac{MAX_I^2}{MSE}\right)
\end{equation}
where $MAX_I$ represents the maximum possible pixel intensity value of the image.

Structural Similarity Index Measure (SSIM) assesses perceived quality by jointly comparing luminance, contrast, and structure between two images, and therefore correlates better with human perception than MSE or PSNR.

\begin{equation}
SSIM(x,y) = \frac{(2\mu_x\mu_y + C_1)(2\sigma_{xy} + C_2)}
{(\mu_x^2 + \mu_y^2 + C_1)(\sigma_x^2 + \sigma_y^2 + C_2)}
\end{equation}

where $\mu_x$ and $\mu_y$ represent the mean intensities of the images, $\sigma_x^2$ and $\sigma_y^2$ represent variances, $\sigma_{xy}$ represents covariance between the two images, and $C_1$ and $C_2$ are small constants used for numerical stability.

Entropy measures the distribution of gray levels. It is reported descriptively: a higher value can reflect either image detail or residual noise, so it is not by itself evidence of better denoising.

\begin{equation}
H = -\sum_{i=0}^{L-1} p_i \log_2 p_i
\end{equation}
where $p_i$ represents the probability of occurrence of intensity level $i$, and $L$ represents the total number of gray levels in the image.
\section{Results}
The proposed Hybrid Quantum Model was compared with classical filtering techniques and a CNN baseline. Tables~\ref{tab_psnr},~\ref{tab_ssim},~\ref{tab_entropy} show the detailed results of PSNR, SSIM and entropy for all 10 images. 
Table~\ref{tab_results} compares the average performance of the evaluated methods across ten test images. Among the classical filters (Gaussian, Median, Bilateral) and the CNN baseline, the proposed Hybrid Quantum Model (HQM) achieves the highest PSNR value of 25.112 and SSIM value of 0.691. 
The proposed HQM attains these scores while operating on only six simulated qubits. Higher PSNR indicates better reconstruction quality, while higher SSIM demonstrates improved preservation of structural information in the denoised images.

Among the classical filters, bilateral filtering performs better than Gaussian and median filtering with a mean PSNR of 24.450 and mean SSIM of 0.668. Gaussian filtering obtains a mean PSNR of 23.388. The CNN baseline obtains a mean SSIM of 0.595, below Gaussian filtering, bilateral filtering, and the HQM.

The proposed model does not achieve the highest SSIM for every test image. It outperforms all baselines clearly on images 9 and 10 (the brain MRI and the NASA-labelled Earth image), with SSIM 0.864 and 0.831, while bilateral filtering has higher SSIM on images 4--8, including differences of 0.121 and 0.097 on images 5 and 6. The higher mean SSIM is therefore not a consistent advantage across images; the roles of image content, learned features and the bilateral post-filter would require ablation experiments.

\subsection{Extended Evaluation on BSD68}\label{sec_bsd68}
To address the limited size of the ten-image study, a second evaluation used the BSD68 test set distributed with DnCNN~\cite{zhang2017dncnn} (68 images). The first 54 images in filename order were used to train a freshly initialized model for 15~epochs (Adam, learning rate 0.001, ReduceLROnPlateau schedule, gradient clipping 1.0, 80 patches per image and noise level), and the remaining 14 images were used for testing. This is a custom 54/14 split, not the standard protocol in which all 68 images form the test set. The bilateral post-filter strength was adapted to the noise level ($\sigma$ 35, 50 and 75 for the three levels). Gaussian, median and bilateral filtering, NLM ($h$ set to the noise level) and BM3D ($\sigma_{\mathrm{psd}}$ set to the noise level) were run on the same noisy images. Table~\ref{tab_bsd68} reports the mean and standard deviation across the 14 test images. A single noise realization was used per image and noise level, so the spread reflects image-to-image variation, not repeated trials.

BM3D achieves the strongest results at every noise level, followed by NLM. The HQM ranks third at $\sigma=15$ (27.68~dB, SSIM 0.815) and $\sigma=25$ (25.22~dB, SSIM 0.716, close to NLM's 25.27~dB and 0.724), ahead of the classical filters, but drops to fifth at $\sigma=50$ (20.69~dB, SSIM 0.489), behind BM3D, Gaussian filtering, NLM and bilateral filtering. Mean inference time per $128\times128$ image on the same machine was 5.5--6.0~s for the HQM versus 1.8--2.2~s for BM3D and under 0.1~s for the filter baselines. The HQM is therefore competitive with classical filters at low and moderate noise but does not match established non-local methods. A training variant with data augmentation scored lower than the reported pipeline at every noise level under the same protocol (mean PSNR 23.66, 22.54 and 19.72~dB, below all baselines) and is not reported further.

\begin{table*}[t]
\caption{Extended evaluation on 14 held-out BSD68 images: mean $\pm$ standard deviation of PSNR (dB) and SSIM across the test images.}
\label{tab_bsd68}
\centering
\footnotesize
\setlength{\tabcolsep}{3.5pt}
\begin{tabular}{|c|c|c|c|c|c|c|}
\hline
Method & PSNR $\sigma{=}15$ & SSIM $\sigma{=}15$ & PSNR $\sigma{=}25$ & SSIM $\sigma{=}25$ & PSNR $\sigma{=}50$ & SSIM $\sigma{=}50$ \\
\hline
Gaussian & $24.350{\pm}3.787$ & $0.711{\pm}0.072$ & $23.623{\pm}3.223$ & $0.648{\pm}0.040$ & $21.540{\pm}2.195$ & $0.502{\pm}0.060$ \\
Median & $24.210{\pm}3.547$ & $0.665{\pm}0.056$ & $22.839{\pm}2.640$ & $0.569{\pm}0.057$ & $19.759{\pm}1.418$ & $0.397{\pm}0.099$ \\
Bilateral & $24.810{\pm}3.372$ & $0.705{\pm}0.096$ & $24.479{\pm}3.075$ & $0.690{\pm}0.064$ & $21.064{\pm}1.377$ & $0.489{\pm}0.118$ \\
NLM & $28.027{\pm}2.450$ & $0.846{\pm}0.058$ & $25.272{\pm}2.817$ & $0.724{\pm}0.087$ & $21.426{\pm}3.160$ & $0.507{\pm}0.123$ \\
BM3D & $28.986{\pm}2.734$ & $0.870{\pm}0.051$ & $26.268{\pm}2.998$ & $0.780{\pm}0.069$ & $23.109{\pm}3.200$ & $0.622{\pm}0.106$ \\
Proposed HQM & $27.675{\pm}2.193$ & $0.815{\pm}0.048$ & $25.224{\pm}1.863$ & $0.716{\pm}0.091$ & $20.694{\pm}0.926$ & $0.489{\pm}0.145$ \\
\hline
\end{tabular}
\end{table*}

\begin{table}[htbp]
\caption{PSNR values for 10 test images across five denoising techniques}
\centering
\footnotesize
\setlength{\tabcolsep}{3pt}
\begin{tabular}{|c|c|c|c|c|c|}
\hline
Image & Gaussian & Median & Bilateral & CNN & Quantum \\
\hline
1 & 23.34 & 21.92 & 23.95 & 23.45 & 23.90 \\
2 & 24.20 & 23.74 & 24.09 & 21.47 & 24.76 \\
3 & 22.57 & 21.81 & 23.27 & 21.74 & 23.93 \\
4 & 23.53 & 24.00 & 25.02 & 22.24 & 25.35 \\
5 & 26.12 & 25.28 & 27.31 & 25.51 & 27.31 \\
6 & 26.38 & 25.60 & 27.11 & 22.73 & 27.44 \\
7 & 20.57 & 20.21 & 23.01 & 20.39 & 24.14 \\
8 & 25.93 & 24.30 & 26.88 & 25.11 & 26.52 \\
9 & 21.35 & 22.86 & 22.33 & 18.09 & 25.00 \\
10 & 19.87 & 20.02 & 21.53 & 17.82 & 22.77 \\
\hline
\end{tabular}
\label{tab_psnr}
\end{table}

\begin{table}[htbp]
\caption{SSIM values for 10 test images across five denoising techniques}
\centering
\footnotesize
\setlength{\tabcolsep}{3pt}
\begin{tabular}{|c|c|c|c|c|c|}
\hline
Image & Gaussian & Median & Bilateral & CNN & Quantum \\
\hline
1 & 0.499 & 0.390 & 0.559 & 0.530 & 0.566 \\
2 & 0.670 & 0.613 & 0.658 & 0.605 & 0.632 \\
3 & 0.630 & 0.570 & 0.632 & 0.624 & 0.688 \\
4 & 0.694 & 0.597 & 0.739 & 0.709 & 0.708 \\
5 & 0.598 & 0.447 & 0.710 & 0.581 & 0.589 \\
6 & 0.680 & 0.541 & 0.770 & 0.658 & 0.673 \\
7 & 0.618 & 0.540 & 0.747 & 0.654 & 0.729 \\
8 & 0.642 & 0.492 & 0.739 & 0.643 & 0.630 \\
9 & 0.594 & 0.678 & 0.598 & 0.504 & 0.864 \\
10 & 0.450 & 0.541 & 0.527 & 0.438 & 0.831 \\
\hline
\end{tabular}
\label{tab_ssim}
\end{table}

\begin{table}[htbp]
\caption{Entropy values for 10 test images across five denoising techniques}
\centering
\footnotesize
\setlength{\tabcolsep}{3pt}
\begin{tabular}{|c|c|c|c|c|c|}
\hline
Image & Gaussian & Median & Bilateral & CNN & Quantum \\
\hline
1 & 6.92 & 7.02 & 6.87 & 6.77 & 7.00 \\
2 & 7.07 & 7.08 & 6.96 & 6.65 & 6.73 \\
3 & 6.96 & 7.12 & 6.88 & 6.81 & 7.04 \\
4 & 7.07 & 7.14 & 6.93 & 6.66 & 7.26 \\
5 & 6.34 & 6.45 & 6.22 & 6.40 & 6.37 \\
6 & 7.51 & 7.56 & 7.44 & 7.23 & 7.52 \\
7 & 7.32 & 7.38 & 7.23 & 7.14 & 7.33 \\
8 & 7.08 & 7.20 & 6.98 & 6.98 & 7.13 \\
9 & 6.98 & 6.50 & 6.73 & 6.29 & 5.85 \\
10 & 6.93 & 6.51 & 6.69 & 6.45 & 5.93 \\
\hline
\end{tabular}
\label{tab_entropy}
\end{table}

\color{black}

\begin{table}[htbp]
\caption{Average PSNR, SSIM, and entropy values across 10 test images for five denoising techniques}
\centering
\footnotesize
\setlength{\tabcolsep}{3pt}
\begin{tabular}{|c|c|c|c|}
\hline
Method & PSNR & SSIM & Entropy \\
\hline
Gaussian & 23.388 & 0.608 & 7.018 \\ 
Median & 22.975 & 0.541 & 6.996 \\
Bilateral & 24.450 & 0.668 & 6.894 \\ 
CNN & 21.855 & 0.595 & 6.737 \\
Proposed HQM & 25.112 & 0.691 & 6.817 \\
\hline
\end{tabular}
\label{tab_results}
\end{table}


\subsection{Discussion: Advantages and Limitations}
The HQM obtains the highest mean PSNR and SSIM among the five evaluated methods on this ten-image set. It combines trainable classical layers, a six-qubit simulated circuit, residual prediction, overlapping reconstruction, and bilateral post-filtering. Consequently, the reported differences cannot be attributed to a quantum advantage without matched classical and post-filter ablations.

Both learned models were trained on the same ten clean images used for evaluation, with newly generated Gaussian noise. There is no held-out image split in this experiment. The results therefore measure denoising of known image content with new noise, not generalization to unseen images. The small heterogeneous image set, one noise level, and a single seeded run further limit the conclusions. No statistical significance, runtime advantage, hardware performance, or standard-benchmark SOTA claim is established.

Only one six-qubit, two-layer circuit configuration was evaluated; ablations of the quantum block and bilateral post-filter and stronger learned baselines remain future work.

\section{Conclusion}
This paper presented a hybrid quantum--classical denoising pipeline with $8\times8$ patches, a six-qubit variational circuit, residual noise prediction, and bilateral post-filtering. On the ten-image, same-content evaluation at noise level $\sigma=0.1$, the complete pipeline obtained mean PSNR 25.112 and SSIM 0.691, the highest means among the five methods evaluated. Per-image SSIM was mixed, and the experiment does not establish generalization or quantum advantage. An extended evaluation on 14 held-out BSD68 images placed the pipeline third behind BM3D and NLM at low and moderate noise and fifth at the highest noise level, confirming its proof-of-concept status. Future work should use the full standard BSD68 protocol, repeated noise trials, and component ablations.

\section*{Acknowledgment}

The authors are thankful to the institutions (Amazon Inc., Thapar Institute of Engineering and Technology, and Maharaja Ranjit Singh Punjab Technical University) for their support. 

{\footnotesize
\bibliographystyle{ieeetr}
\bibliography{ref}
}

\end{document}